\documentclass[letterpaper, 9 pt, conference]{ieeeconf}  % Comment this line out
\IEEEoverridecommandlockouts                              % This command is only
\usepackage[utf8]{inputenc}
\usepackage[T1]{fontenc}
\usepackage{algorithm}
\usepackage{hyperref}
\usepackage{algorithmic}
\usepackage{amsmath}
\usepackage{enumerate}
\usepackage{bbm}
\usepackage{amssymb}
\usepackage[usenames,dvipsnames]{color}
\usepackage{listings}
\usepackage{geometry}
\usepackage{environ}         % provides \BODY
\usepackage{etoolbox}        % provides \ifdimcomp
\usepackage{graphicx}        % provides \resizebox
\usepackage{amsthm}

\newlength{\myl}
\let\origequation=\equation
\let\origendequation=\endequation

\RenewEnviron{equation}{
  \settowidth{\myl}{$\BODY$}                       % calculate width and save as \myl
  \origequation
  \ifdimcomp{\the\linewidth}{>}{\the\myl}
  {\ensuremath{\BODY}}                             % True
  {\resizebox{\linewidth}{!}{\ensuremath{\BODY}}}  % False
  \origendequation
}

\newtheoremstyle{indented}
  {3pt}% space before
  {3pt}% space after
  {\addtolength{\linewidth}{1.5em}
   \addtolength{\linewidth}{-2.5em}
   \parshape 1 1em \linewidth}% body font
  {}% indent
  {\bfseries}% header font
  {.}% punctuation
  {.5em}% after theorem header
  {}% header specification (empty for default)
\makeatother

\theoremstyle{indented}
\newtheorem{theorem}{Theorem}

\title{\LARGE \bf
Optimal Control of Agent-Based Dynamics under Deep Galerkin Feedback Laws
}

\author{Frederik Kelbel$^{1}$ %, Dante Kalise$^{2}$, Grigorios Pavliotis$^{3}$ % <-this % stops a space
\thanks{*This work was not supported by any organization}% <-this % stops a space
\thanks{$^{1}$F. Kelbel is with Department of Mathematics,
        Imperial College London, South Kensington Campus, London SW7 2AZ, UK
        {\tt\small fjk20@ic.ac.uk}}}
\begin{document}
\newgeometry{margin=2cm,textwidth=\paperwidth}

\maketitle
\thispagestyle{empty}
\pagestyle{empty}

%%%%%%%%%%%%%%%%%%%%%%%%%%%%%%%%%%%%%%%%%%%%%%%%%%%%%%%%%%%%%%%%%%%%%%%%%%%%%%%%
\begin{abstract}
Ever since the concepts of dynamic programming were introduced, one of the most difficult challenges has been to adequately address high-dimensional control problems. With growing dimensionality, the utilisation of Deep Neural Networks promises to circumvent the issue of an otherwise exponentially increasing complexity. The paper specifically investigates the sampling issues the Deep Galerkin Method is subjected to. It proposes a drift relaxation-based sampling approach to alleviate the symptoms of high-variance policy approximations. This is validated on mean-field control problems; namely, the variations of the opinion dynamics presented by the Sznajd and the Hegselmann-Krause model. The resulting policies induce a significant cost reduction over manually optimised control functions and show improvements on the Linear-Quadratic Regulator problem over the Deep FBSDE approach.
\end{abstract}

%%%%%%%%%%%%%%%%%%%%%%%%%%%%%%%%%%%%%%%%%%%%%%%%%%%%%%%%%%%%%%%%%%%%%%%%%%%%%%%%
\section{INTRODUCTION}
Deep Learning has been shown to be an effective tool in finding the solution of high-dimensional PDEs \cite{sirignano, exarchos, weinan-e, weinan-e2, raissi, al-aradi}. With the application of the Deep Galerkin Method to stochastic optimal control problems, the appropriate sampling of batches on domain and boundary becomes a major challenge. This is especially the case when considering the dynamics of interacting agents. We show how the sampling technique may prohibit the Deep Galerkin loss from converging to zero and propose a simple algorithm to alleviate this issue. The approach is evaluated on the controlled consensus dynamics represented by the Szanjd model \cite{sznajd} and the Hegselmann-Krause Model \cite{hegselmann}. \\ \\ 
In contrast to open-loop control, feedback control laws do not depend on a specific initial condition and are significantly more robust to perturbations. Optimal feedback controllers can be composed from the solution of the respective Hamilton-Jacobi-Bellman equation. The solution of this partial differential equation, however, becomes intractable in higher-dimensional problems. In the case of unconstrained linear control systems with quadratic cost, the Hamilton-Jacobi-Bellman equation reduces to an Algebraic Riccati equation. This has been extensively studied in \cite{kalman} early on in 1960. The first significant approaches on nonlinear control systems with a scalar control variable were made in \cite{albrekht}. The authors' method involved a power series expansion of the involved terms around the origin. They inserted these back into the Hamilton-Jacobi-Bellman equation and collected expressions of similar order. He obtained a sequence of algebraic equations part of which conveniently reduce to the Riccati equation. Each of these expressions can be solved. The solution is a local solution in a neighbourhood around the origin. More recent efforts focused on semi-Lagrangian schemes \cite{bokanowski} \cite{falcone}. These work similarly to Finite Difference Methods. However, they also employ an interpolation scheme for the region surrounding grid points. Like other grid-based approaches, these methods do not scale well to higher dimensions. The authors of \cite{alla} alleviate this issue by coupling grid-based discretisations of low-dimensional Hamilton-Jacobi-Bellman equations. They base this method on the concept of proper orthogonal decomposition; a technique that is well known from computational fluid dynamics. However, the quality of these schemes has been shown to deteriorate in highly nonlinear or advection affected settings such as presented in \cite{kalise2}. In \cite{kang}, they compute approximate solutions to Hamilton-Jacobi-Bellman equations by combining the method of characteristics with sparse space discretisations. This works similarly to the semi-Lagrangian schemes. First, they solve on a very coarse grid using the characteristic equations, then, polynomial interpolations are used to obtain approximations at arbitrary points. This approach is causality-free, i.e. it does not directly depend on the density of the grid and is, therefore, more suitable for high-dimensional problems. Alternative implementations based on tensor decomposition have also been proven successful in tackling dimensionality issues \cite{stefansson}. \cite{dolgov} extended such framework to fully nonlinear, first-order, stationary Hamilton-Jacobi-Bellman PDEs. Nevertheless, the curse of dimensionality remains a challenge. There are several causality-free deep learning algorithms that are applicable to high-\hspace{0pt}dimensional stochastic optimal control.
The idea of using data-driven value function approximations is not new. An early record of this is \cite{munos} in which a neural network was utilised to model the solution to the Hamilton-Jacobi-Bellman equation associated with the control of a car in a one-dimensional landscape. FBSDE-based methods such as in \cite{raissi}, \cite{exarchos}, or \cite{weinan-e2} solve the associated system of Forward-Backward SDEs. This is realised by integration over a time-discretised domain of path realisations. Another idea relies on the minimisation of the residuals of the Hamilton-Jacobi-Bellman PDE \cite{al-aradi} \cite{sirignano} \cite{nguyen}. This concept is termed the Deep Galerkin Method and forms the foundation of this paper. Opinion dynamics models such as in \cite{sznajd} and \cite{hegselmann} are governed by interacting diffusion processes. Their models are the basis to the realisation of the endeavour to enforce coherent behaviours in large populations. The control of these is manifested as a mean-field control problem. \cite{kalise} approximates the optimal policy via a hierarchy of suboptimal controls. Instead, the proposal is, here, to employ a Deep Galerkin approach. \\ \\
The paper is structured as follows. In Section \ref{sec:background}, we introduce the relevant background later sections build on. This includes information regarding the general concepts in Stochastic Optimal Control, the Hamilton-Jacobi-Bellman equation, and the Deep Galerkin Method. Subsection \ref{sec:agent_based_dynamics} goes more into detail about the type of problems considered, while Section \ref{sec:sampling_problem} exemplifies the sampling issues that appear with the Deep Galerkin Method when considering the optimal control of interacting, stochastic agents. The proposed methodology is evaluated in the last section. The code repository is available at {\color{blue}\href{https://github.com/FreditorK/Optimal-Control-of-Agent-Based-Dynamics}{https://github.com/FreditorK/Optimal-Control-of-Agent-Based-Dynamics}}.

\section{BACKGROUND}\label{sec:background}

\subsection{Stochastic Optimal Control}
We study a finite horizon problem over the interval $[t, T]$ subjected to nonlinear stochastic dynamics as represented by an Itô process. We denote by $(\Omega, \mathcal{F}, \{\mathcal{F}_t\}_{t \geq 0}, \mathbb{P})$ a filtered probability space and by $\mathbb{A} \subseteq L^2([t, T] \times \Omega; \mathbb{P})$ the space of admissible control functions. We consider the optimisation problem

\begin{align}
\nonumber \inf_{\upsilon \in \mathbb{A}} \; & \mathbb{E}^\mathbb{P}\Big[ \int_t^T F(s, X_s, \upsilon(s, X_s)) ds + G(X_T) \; \Big| \; X_t = x \Big] \\
\nonumber \text{s.t. } &d X_s = \mu(s, X_s, \upsilon(s, X_s)) ds + \sigma(s, X_s, \upsilon(s, X_s)) d W_s \\
&X_0 \sim \nu,
\label{eq:optimisation_problem}
\end{align}
where $X_s \in \Omega \subseteq \mathbb{R}^n$ is an $\{\mathcal{F}_s\}_{s \geq 0}$-adapted Itô process, $\mu: \mathbb{R} \times \mathbb{R}^n \times \mathbb{R} \mapsto \mathbb{R}^n$, $\sigma: \mathbb{R}\times\mathbb{R}^n \times \mathbb{R} \mapsto \mathbb{R}^{n \times m}$, and $W_s$ is an $m$-dimensional Brownian motion. The initial distribution of the process $\{X_s\}_{s \geq 0}$ is denoted by $\nu$. Additionally, the objective is specified by a cost function $F: \mathbb{R} \times \mathbb{R}^n \times \mathbb{R} \mapsto \mathbb{R}^+$ and a terminal cost $G: \mathbb{R}^n \mapsto \mathbb{R}^+$. These are assumed to be non-negative bounded from below and continuous. 
\subsection{The Hamilton-Jacobi-Bellman Equation}
We denote by $J: \mathbb{R} \times \mathbb{R}^n \mapsto \mathbb{R}$ the optimal value function such that
\begin{align*}
    J(t, x) = \inf_{\upsilon \in \mathbb{A}} \; & \mathbb{E}^\mathbb{P}\Big[ \int_t^T F(s, X_s, \upsilon(s, X_s)) ds + G(X_T) \; \Big| \; X_t = x \Big]
\end{align*}
By Bellman's Optimality Principle, the value function $J$ satisfies the Hamilton-Jacobi-Bellman equation
\begin{align*}
\begin{cases}
    0 = \partial_t J(t, x) + \underset{u \in \mathbb{R}}{\min} \{\mathcal{L}^u J(t, x) + F(t, x, u) \} \\
    J(T, x) = G(x)
\end{cases}, \\
\text{where } \mathcal{L}^u := \mu(s, X_s, u) \cdot \nabla + \frac{1}{2} \sigma(s, X_s, u) \sigma(s, X_s, u)^T : D^2
\end{align*}
with $u$ being substituted in as the optimal control value closed-loop control process within $\mathbb{R}$. $D^2$ represents the Hessian operator here. Under the assumption that the Hamiltonian $H(t, x, u) = \mathcal{L}^u J(t, x) + F(t, x, u)$ is differentiable and the space of admissible control functions is unconstrained, the optimal control process can often times be recovered by solving $\frac{d}{du} H(t, x, u) = 0$. The examples in this paper make use of this property.

\subsection{The Deep Galerkin Method}
Let $u$ represent the optimal control signal at time $t$. The Deep Galerkin Method \cite{sirignano} aims to minimise the residuals of the differential terms of the $\theta$-parametrised neural network $J_\theta$ to the solution of the Hamilton-Jacobi-Bellman equation $J$. Given an equation
\begin{align*}
\begin{cases}
    0 = \partial_t J(t, x) + \mathcal{L}^{u} J(t, x) + F(t, x, u) \\
    J(T, x) = G(x)
\end{cases}.
\end{align*}
we minimise the error $\mathcal{E}(J_\theta; [t, T] \times \Omega)$ composed of
\begin{align}
    \nonumber\mathcal{E}(J_\theta; [t, T] \times \Omega) =& w_1 ||(\partial_s + \mathcal{L}^{u}) (J_\theta - J)||_{L^2([t, T] \times \Omega; \nu_1)} \\ \nonumber
    & + w_2 ||J_\theta(T, \cdot) - G||_{L^2(\Omega; \nu_2)} \\ \nonumber
    =& w_1 ||(\partial_s + \mathcal{L}^{u}) J_\theta + F||_{L^2([t, T] \times \Omega; \nu_1)} \\
    & + w_2||J_\theta(T, \cdot) - G||_{L^2(\Omega; \nu_2)}.
    \label{eq:dgm_obj}
\end{align}
The respective norms are weighted with the scalars $w_1, w_2 \in \mathbb{R}$ and are approximated by taking the mean of the squared residual over $\nu_1$- and $\nu_2$-sampled batches. However, as we will see in Theorem \ref{th:bound}, we can issue a recommendation on the allocation of the weightings. This is done iteratively along with Gradient Descent as formulated in Algorithm \ref{alg:DGM}. A batch purposed for the first term in the loss is denoted by $\mathcal{B}_{\nu_1}$ while the other batch is defined with $\mathcal{B}_{\nu_2}$.

\begin{algorithm} 
\caption{Deep Galerkin Approximation}
\textbf{Input:} \\
Batch sizes $N, M \in \mathbb{N}$, \\
Parameters $\theta$, \\
Learning rate $\alpha$, \\
Error threshold $\varepsilon > 0$, \\
$(\nu_1, \nu_2)$ some given distributions \\
\textbf{Output:} Parameters $\theta$ 
\begin{algorithmic}
    \WHILE{$\mathcal{E}(J_\theta; \mathcal{B}_{\nu_1}, \mathcal{B}_{\nu_2}) > \varepsilon$}
        \STATE $\mathcal{B}_{\nu_1} = (\tau_i, x_i)_{i=0}^N \sim \nu_1[[t, T] \times \Omega]$ \STATE $\mathcal{B}_{\nu_2} = (z_i)_{i=0}^M \sim \nu_2[\Omega]$
        \STATE Propagate the samples through $J_\theta$ and compute $\mathcal{E}(J_\theta; \mathcal{B}_{\nu_1}, \mathcal{B}_{\nu_2})$
        \STATE $\theta \leftarrow \theta - \alpha \nabla_\theta \mathcal{E}(J_\theta; \mathcal{B}_{\nu_1}, \mathcal{B}_{\nu_2})$
    \ENDWHILE
\end{algorithmic}
\label{alg:DGM}
\end{algorithm}
\vspace{0.1mm}

\subsection{Agent-Based Dynamics} \label{sec:agent_based_dynamics}
Imagine a system of interacting agents in which we seek to endorse some collective behaviour. The dynamics of agent $i$ in such a system is represented by 
\begin{align*}
    d X^{(i)}_s =& \mu(t, X^{(i)}_t, X_t, \upsilon_i(t, X_t)) ds \\
    &+ \sigma(t, X^{(i)}_t, X_t, \upsilon_i(t, X_t)) d W_s,
\end{align*}
where $X_t$ is the collective state process. Fix the number of agents at $n$. To describe the behaviour of large populations of agents, we make use of the representation capabilities in Mean Field Control Theory. The cost is formulated in terms of the discrepancies in the law of the process $\mathbb{P}$ approximated by the empirical measure $\mathbb{P}_n = \frac{1}{n} \sum_{i=1}^n \delta_{X^{(i)}}$, i.e. the average of the probability point masses. For this, we use the squared Wasserstein metric $\mathcal{W}^2_{\Omega, 2}$ and a cost functional $\psi$. The interaction of the agents is restricted to the drift term and realised via the interaction kernel $P$. Crowd control of this form has been studied in \cite{albi, kalise, degond}. Interpreted in the mean-field sense, we optimise over measure flows on $[t, T]$ and the crowd control problem can be generalised to be of the form:\\
\begin{align}
\begin{array}{rll}
     \underset{\{\upsilon_i\}_{i \geq 1} \subseteq \mathbb{A}}{\min} & \mathbb{E}^\mathbb{P}\Big[ \frac{1}{2n} \int_t^T \mathcal{W}^2_{\Omega, 2}(X_s, x_d)+  \sum_{i=1}^n \psi(\upsilon_i(s, X_s)) ds \\ & \hspace{5mm} +  \frac{1}{2n} \mathcal{W}^2_{\Omega, 2}(X_T, x_d)\; \Big| \; X_t = x\Big] \\
     \text{s.t.} & d X^{(i)}_s = \frac{1}{n} \sum_{j=1}^n P(X^{(i)}_s, X^{(j)}_s)(X^{(j)}_s - X^{(i)}_s) \\ & \hspace{12mm} + \upsilon_i(s, X_s) ds + \sigma d W_s, \text{for } 1 \leq i \leq n \\
     & X_0^{(i)} \sim \nu.
\end{array}
\label{eq:agent_dynamics}
\end{align}
We define $x_d$ to be the target measure of the optimisation problem, i.e. a desirable state within $[t, T]$ and at terminal time. Notice that the squared Wasserstein metric simplifies to the squared $l_2$-norm on finite vector spaces with $x_d$'s entries being single valued.

\section{THE SAMPLING PROBLEM}\label{sec:sampling_problem}
The problem described in this section is two-fold and is concerned with error term selected in Equation \ref{eq:dgm_obj}, more specifically the choice of measure from which to sample. Firstly, let's establish the relationship between the minimisation of the HJB-residual and the regression of the parameterised network to the true value function:
\begin{align*}
   ||\underbrace{J_\theta - J}_{\delta_\theta}||_{L^2(\Omega; \mathbb{P})}. 
\end{align*}
The relation becomes clear in the statement of Theorem \ref{th:bound}.
\begin{theorem}\label{th:bound}
Let $(\Omega, \mathcal{F}, \mathbb{P})$ be a probability space, then the $L^2$-error of the value function $J_\theta$ to the true value function $J$ at time $t$ is bounded by above by the residuals of the Hamilton-Jacobi-Bellman equation:
\begin{align*}
 \sqrt{T-t} & \big|\big|(\partial_t + \mathcal{L}^{u}) J_\theta + F(\cdot, \cdot, u) \big|\big|_{L^2([t, T] \times \Omega; \mathbb{P})} \\
 + &||J_\theta(T, X_T) - G(X_T)||_{L^2(\Omega; \mathbb{P})} \\
 &\geq ||J_\theta(t, \cdot) - J(t, \cdot)||_{L^2(\Omega; \mathbb{P})} 
\end{align*}
Proof: See appendix.
\end{theorem}
\noindent The DGM-loss gives an upper bound for the $L^2$-error of the parameterised value function for any time $t$. Or formulated the other way around, the error formulated as a regression problem gives a lower bound to the DGM-loss. We will use this to show, that the only sensible sampling measure is the law of the stochastic process. For this, recall that the solution to the Hamilton-Jacobi-Bellman PDE is the conditional expectation:
\begin{align*}
    Y_t = J(t, x) = \mathbb{E}^{\mathbb{P}}\Big[ \int_t^T F(s, X_s, u_s) ds + G(X_T) \;\Big|\; X_t = x \Big].
\end{align*}
This causes two potential issues. The conditional expectation is only unique up to measure zero conditioning, i.e. it becomes a problem under the $\nu_1$-measure if a sample $(t, x)$ appears that is $\mathbb{P}$-measure zero. This is thought to introduce noise. There is however another argument to be made; even if the conditional expectation is well-defined. We manifest the following result:
\begin{theorem}\label{th:bound_2}
Let $(\Omega, \mathcal{F}, \mathbb{P})$ be a probability space and $\sigma_{\nu_1}(Z_t)$ be the sigma algebra generated by the $\nu_1$-random variable $Z_t$. Let $Y_t$ be well-defined on $\sigma_{\nu_1}(Z_t)$. Further, let the running cost and terminal cost be bounded by below with $F\geq B_F \in \mathbb{R}^+$ and $G\geq B_G \in \mathbb{R}^+$. Then, the DGM-loss is bounded below by 
\begin{align*}
\mathcal{E}(J_\theta;& [t, T] \times \Omega)  \\
    &\geq ((T-t) B_F + B_G)^2 ||\mathbb{P}(X_t \in \mathcal{F}_t \setminus \sigma_{\nu_1}(Z_t))||^2_{L^2(\Omega; \mathbb{P})}
\end{align*}
Proof: See appendix.
\end{theorem}
\noindent The magnitude of the lower bound for the loss depends on $\mathbb{P}(X_t \in \mathcal{F}_t \setminus \sigma(Z_t))$. Independently from the training time, the error will not converge to zero unless the sampling measure produces the filtration. It is quite intuitive. The closer the samples resemble the process, the better the approximation. \\ \\
The proposal is to construct samples from $\mathbb{P}$ during the training based on an approximate optimal policy. Let $\xi = \{\xi_i\}_{i=1, ..., m} \sim \mathcal{N}(0, 1)$, where $\sigma$ is an $n \times m$ matrix. Under the Euler-Maruyama scheme, and discretisation $\Delta t$ this becomes
\begin{align*}
    X^{(i)}_{t+\Delta t} =& \text{ SDE}(t, X_t, u_t) = X^{(i)}_{t} + \mu(t, X_t, u_t) \Delta t + \sigma \sqrt{\Delta t} \xi, \\ & i = 1, ..., n, \quad X_0 = x.
\end{align*}

\begin{figure}[!htb]
    \centering
    \includegraphics[width=0.9\linewidth, trim={0 0 0 3cm}, clip]{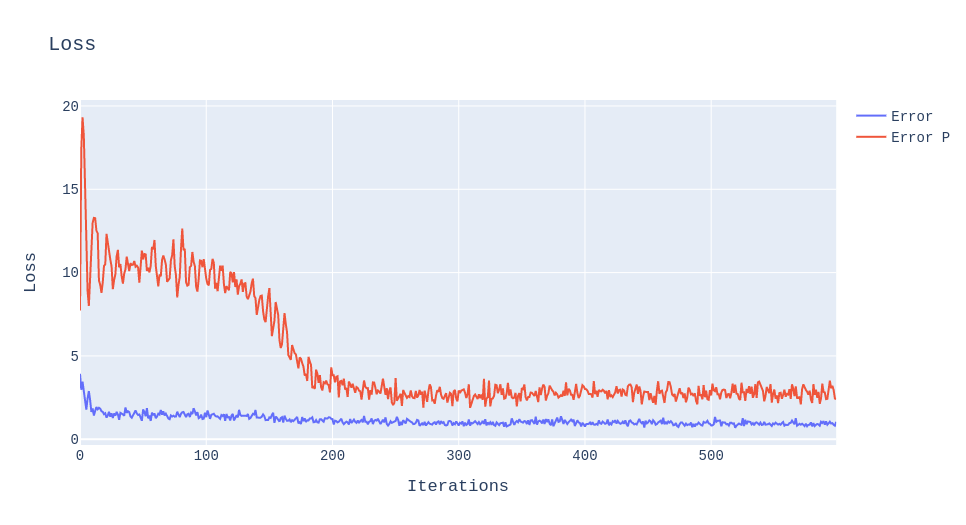}
    \caption{$\mathcal{E}(J_\theta; \mathcal{B}_{\pi_U})$ (Error) and $\mathcal{E}(J_\theta; \mathcal{B}_\mathbb{P})$ (Error P) determined w.r.t. a model trained on a uniform sampling scheme, i.e. samples were drawn from a uniform measure $\pi_U$ during training. Latter error was computed on a batch produced by the Euler-Mayurama scheme using the model's policy. The $\mathcal{E}(J_\theta; \mathcal{B}_\mathbb{P})$-error shows a high variance and poor convergence properties at a high magnitude. $\mathcal{E}(J_\theta; \mathcal{B}_{\pi_U})$ converges to a value of $0.87$, $\mathcal{E}(J_\theta; \mathcal{B}_\mathbb{P})$ oscillates between $3-4$ (Produced on Sznajd model as per Section \ref{sec:evaluation}). }
    \label{fig:error_sampling_1}
\end{figure}
\begin{figure}[!htb]
    \centering
    \includegraphics[width=0.9\linewidth, trim={0 0 0 3cm}, clip]{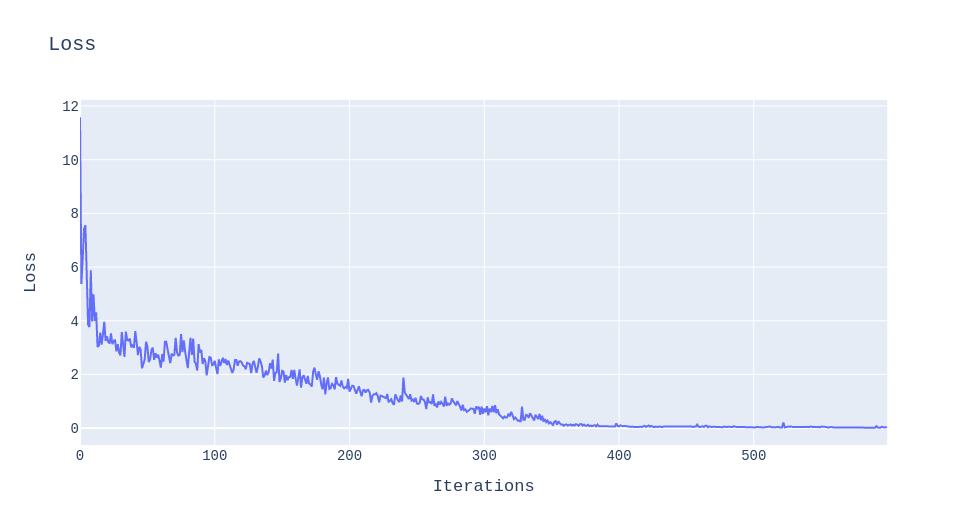}
    \caption{$\mathcal{E}(J_\theta; \mathcal{B}_\mathbb{P})$ trained with a $\mathbb{P}$ sampling scheme as per Algorithm \ref{alg:controlled_path} (Produced on Sznajd model as per Section \ref{sec:evaluation}). The error converges to zero.}
    \label{fig:error_sampling_2}
\end{figure}

\noindent The algorithm works as follows. Initially a batch is sampled from a distribution $\nu(\Omega)$. The time points $\{t_i\}_{i=1}^N$ are sampled uniformly or quasi-uniformly over $[0, T]$. The algorithm starts by sampling from the uncontrolled path space, i.e. with $\alpha=1$ and $\text{SDE}(t_i, x, (1-\alpha)u), \; i = 1, ..., n$. From these fully-relaxed dynamics, the SDE is gradually introduced to the control signal. The variance of the control signal is rather high in the beginning. The rational is that the algorithm reduces the variance until a better approximation of the control function is available. After each propagation, the modulus is taken on the updated time as to maintain a uniform distribution over the time horizon. One sample is generated in $\mathcal{O}(N)$, for a batch size of $N$. The methodology is displayed in Algorithm \ref{alg:controlled_path}. It is straightforward to transform the algorithm into an its off-policy version via the extension with a replay buffer. As seen in Figure \ref{fig:error_sampling_1} uniform sampling provides poor convergence in both the uniform $L^2$-norm and the $L^2$-norm with respect to an approximation of the law of the process.
\begin{algorithm}[!htb]
\caption{Controlled Drift Relaxation}
\textbf{Input:} \\
Batch size $N \in \mathbb{N}$, \\
Sample $(x, t, u) = \{(x_i, t_i, u_i)\}_{i=1}^N, x_i \in \mathbb{R}^n, t_i \in [0, T]$, $u_i \in \mathcal{A}$\\
Learning rate $\beta$, \\
Initial distribution $\nu$, \\
Target measure $x_d$, \\
Relaxation coefficient $\alpha \in [0, 1]$, \\
Stochastic Differential Equation SDE$(\cdot, \cdot, \cdot)$\\
\textbf{Output:} Sample $(x, t)$
\begin{algorithmic}
\WHILE{$\alpha > 0$}
    \FOR{$i = 1...N$}
    \STATE $x^{\text{new}}_i \sim \nu(\Omega)$
    \STATE $x_i \leftarrow [t_i + \Delta t \leq T] \cdot \text{SDE}(t_i, x_i, (1-\alpha)u) + [t_i + \Delta t > T] \cdot x^\text{new}_i$
    \STATE $t_i \leftarrow (t_i + \Delta t) \mod T$
    \ENDFOR
    \STATE $\alpha \leftarrow \alpha - \beta \nabla_\alpha (\alpha^2 + \sum_{i=1...N} \frac{t_i}{T N}\mathcal{W}^2_{\Omega, 2}(x_i, x_d))$
    \STATE $\alpha \leftarrow \max(\min(\alpha, 1), 0)$
\ENDWHILE
\end{algorithmic}
\label{alg:controlled_path}
\end{algorithm}
Using Algorithm \ref{alg:controlled_path} to sample the batches, however, allows the norm with respect to the approximate solution to go to zero. This is displayed in Figure \ref{fig:error_sampling_2}. Note that the algorithm can be extended to provide boundary samples by adding the batch entries that fall within the boundary to a replay buffer from which one can draw uniform samples. The algorithm scales the policy's output. The convergence towards the target measure is improved on. 

\section{NUMERICAL EVALUATION}\label{sec:evaluation}
For the Deep Galerkin Method we employ a residual neural network with three layers. Each residual layer is a two-layer perceptron with the respective skipping connection. The FBSDE-model uses a simple three-layer perceptron. Both networks are built on SiLU-activations. We optimise using ADAM with an initial learning rate of $10^{-3}$ and a decay of $0.99$ for every ten training iterations in which the loss plateaus. The initial distribution is realised by Algorithm \ref{alg:clustered_sampling}. The sampling algorithm aims to improve the policy's ability to generalise to unknown initial distributions. Let $n$ be the dimension of the underlying diffusion process of the Hamilton-Jacobi-Bellman equation. The algorithm starts by uniformly sampling $n$ points from the domain. Let the resulting set of samples be given by $\{p_j\}_{j=1}^n$. From this set it selects a subset $\{p_{k_j}\}_{j=1}^n$, where $\{ k_j \}_{j=1}^n$ are indices sampled from a truncated normal distribution $\mathcal{N}_T$ over the integer set $\{1, 2, ..., n\}$. The standard deviation of this distribution essentially adjusts the variance in $\{ p_{k_j} \}_{j=1}^n$. A high standard deviation yields more uniformly distributed samples $\{ k_j \}_{j=1}^n$. It regulates the amount of duplicates in $\{p_{k_j}\}_{j=1}^n$. The resulting subset contains the means on which uniform distributions are centred forming balls with maximum width $\epsilon$. For fixed and sufficiently large $\epsilon$ the samples approach a uniform distribution as $\sigma$ goes to infinity. Alternatively, one can also sample $\epsilon$ from an appropriate range of values. Fix $\epsilon$ such that the balls $\cup_{i=1}^n B(p_{i}, \epsilon)$ cover the domain. From each ball $B(p_{k_j}, \epsilon)$, we then draw a point uniformly. The sampling algorithm is shown in Algorithm \ref{alg:clustered_sampling}. This results in a variety of more or less clustered agents at initial time.

\begin{algorithm}[!htb]
\caption{Clustered Sampling}
\textbf{Input:} \\
Batch size $N \in \mathbb{N}$, \\ 
Lower and upper bound $b_l, b_u \in \mathbb{R}$, \\
Standard deviation for truncated normal $\sigma \in (0, \infty)$ \\
Length of partitions $\epsilon \in (0, |b_l| + |b_u|)$ \\
\textbf{Output:} Clustered samples $x = \{x_i\}_{i=1}^N \in [b_l, b_u]^{N \times n}$
\begin{algorithmic}
\FOR{i=1, ..., N}
    \STATE $\{p_j\}_{j=1}^n \sim \mathcal{U}[b_l, b_u]$
    \STATE $\{ k_j \}_{j=1}^n \sim \mathcal{N}_T(0, \sigma)$
    \STATE $m_i \leftarrow \{p_{k_j}\}_{j=1}^n$
    \STATE $a_i \sim \mathcal{U}[0, 1]$
    \STATE $b_i \leftarrow \min(|m_i-b_l|, |m_i-b_u|)$
    \STATE $x_i \leftarrow a_i \cdot \min(b_i, \epsilon) + m_i$
\ENDFOR
\end{algorithmic}
\label{alg:clustered_sampling}
\end{algorithm}

\subsection{Linear Quadratic Regulator}
\begin{figure}[!htb]
    \centering
    \includegraphics[width=0.9\linewidth, trim={0 0 0 1.8cm}, clip]{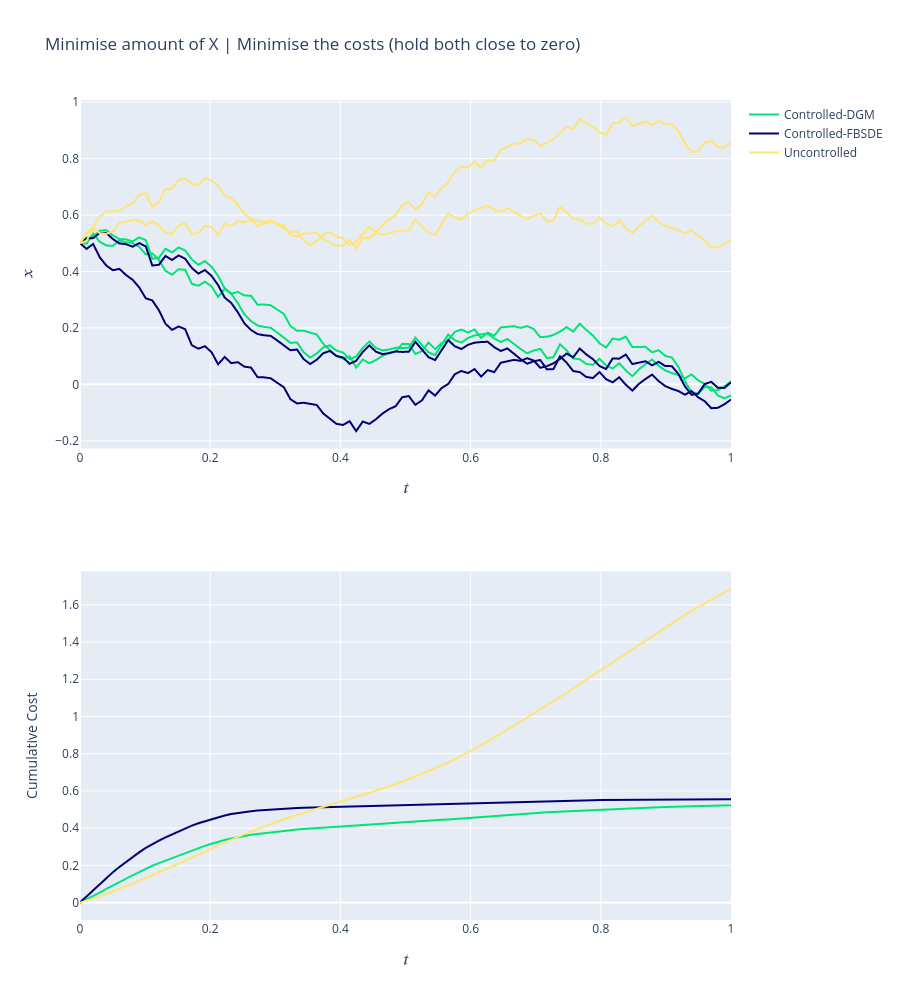}
    \caption{Controlled dynamics in accordance with the Linear Quadratic Regulator and $2$ agents. The interval is discretised with 100 time points. Using a sampler as in Algorithm \ref{alg:controlled_path} the Deep Galerkin Method performs better and at a lower complexity than an FBSDE-approach.}
    \label{fig:lqr}
\end{figure}

This subsection makes comparisons to an FBSDE-based approach as in \cite{exarchos} with the optimal control introduced into the Forward SDE via a change of measure and a DGM-based approach with controlled drift relaxation. We consider a Linear Quadratic Regulator Problem of the following form:
\begin{align}
\begin{array}{rll}
     \underset{\upsilon \subseteq \mathbb{A}}{\min} & \mathbb{E}^\mathbb{P}\Big[ \int_t^T  X^T_s C X_s+ \frac{1}{2} \upsilon(s, X_s) D \upsilon^T(s, X_s) ds \\
     & \hspace{10mm} + X_T^T R X_T \; \Big| \; X_t = x\Big] \\
     \text{s.t.} & dX_s = [H X_s + M \upsilon(s, X_s)] ds + \sigma dW_s, \\
     & X_0 \sim \nu
\end{array},
\end{align}
\noindent where $\sigma = 0.2$ and the matrices are given by
\begin{align*}
    H &= \begin{bmatrix} 0.1 & 0 \\ 0.05 & 0.1 \end{bmatrix}, \;
    M = \begin{bmatrix} 1 & 0 \\ 0 & 1 \end{bmatrix}, \;
    C = \begin{bmatrix} 2 & 0 \\ 0 & 2 \end{bmatrix}, \\
    R &= \begin{bmatrix} 0.1 & 0 \\ 0 & 0.1 \end{bmatrix}, \;
    D = \begin{bmatrix} 0.2 & 0 \\ 0 & 0.2 \end{bmatrix}.
\end{align*}
Both neural networks are chosen to fall into the $91 \; 000 - 92 \; 000$ parameter range. The network used in the FBSDE-model is a simple 3-layer Perceptron instead of the residual model simply for empirical reasons. The number of discretisations steps are deliberately chosen at $100$ as to show the advantage of the DGM Method. At this horizon length, the gradients easily vanish and it also becomes insensible to utilize recurrent neural network structures for Deep FBSDE-models. The complexity of the FBSDE-model is highly dependent on the time discretisation. Both models are trained for $600$ iterations. However, the FBSDE-model takes around $18$ times longer to run than the DGM-model which terminates within $25$ seconds train on a NVIDIA GeForce MX350 with $2$ GB of dedicated GDDR5 memory. The sampler's $\alpha$-parameter is decayed at a rate of $\beta=0.1$. Figure \ref{fig:lqr} shows one path realisation with two agents and their associated cumulative cost over a time horizon of $[0, 1]$. We observe a significant difference between the estimated policies. The approximation the Deep Galerkin Method is much more conservative. It performs slightly better than the FBSDE-equivalent.

\subsection{Sznajd Model}
\begin{figure}[!htb]
    \centering
    \includegraphics[width=0.9\linewidth, trim={0 0 0 1.5cm}, clip]{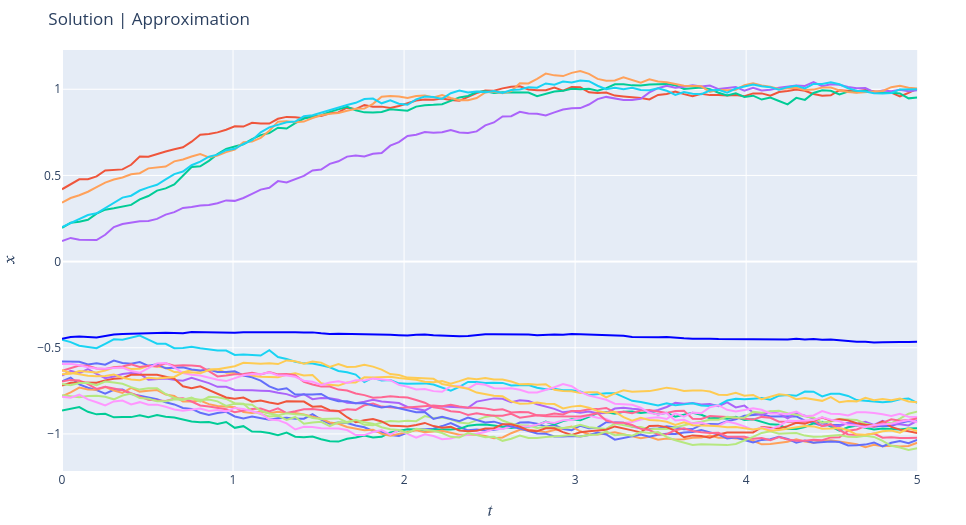}
    \caption{Uncontrolled opinion dynamics in accordance with the Sznajd model and $20$ agents. The interval $[0, 5]$ is discretised with $100$ time points. The problem is specified with $\beta=-3$, $\sigma=0.01$, $\gamma=0.04$, and $\lambda=1$. Each agent is associated with one coloured line. The bold blue line represents the empirical average.}
    \label{fig:uncontrolled_sznajd}
\end{figure}
In this subsection, the $n$-agent system from Section \ref{sec:agent_based_dynamics} evolves accordingly with the model by \cite{sznajd}. The previous opinion dynamics are expanded by an additional drift term. The Sznajd model fixes $\psi(c) = \gamma |c|^2$ and $P(x, y) = \beta (1-x^2)$. It models the propensity of an agent to change their opinion within the domain $\Omega = [-1, 1]$. Towards either boundary the influence of one agent on another agent decreases. For $\beta > 0$, consensus appears naturally and a flocking can be observed. For $\beta < 0$, a polarisation of the agents towards the extrema occurs. This polarisation is depicted in Figure \ref{fig:uncontrolled_sznajd}. Under these circumstances, the problem can be reformulated as:
\begin{align}
\begin{array}{rll}
     \underset{\{\upsilon_i\}_{i \geq 1} \subseteq \mathbb{A}}{\min} & \mathbb{E}^\mathbb{P}\Big[ \int_t^T  \frac{1}{2n} \sum_{i=1}^n \lambda |X^{(i)}_s - x^{(i)}_d|^2 \\ 
     & \hspace{4mm} + \gamma |\upsilon_i(s, X_s)|^2 ds + \frac{\lambda}{2n} ||X_T - x_d||_2^2\; \Big| \; X_t = x\Big] \\
     \text{s.t.} & dX^{(i)}_s = \beta (1-(X^{(i)}_s)^2) (\overline{X}_s - X_s^{(i)}) ds \\
     & \hspace{8mm} + \upsilon_i(s, X_s) ds + \sqrt{2 \sigma} dW^{(i)}_s, \text{for } 1 \leq i \leq n \\
     & X_0 \sim \nu.
\end{array}
\end{align}
\noindent We associate the following Hamilton-Jacobi-Bellman equation to the problem:
\begin{align}
\begin{cases}
    0 = \partial_t J + \frac{\lambda}{2n} |x-x_d|^2 + (\beta (\vec{1}- x \odot x) \odot (\bar{x} \vec{1} - x ))^T \nabla_x J  \\ 
     \hspace{5mm} + \sigma \Delta_x J - \frac{n}{2 \gamma}|\nabla_x J|^2\\
    J(T, x) = \frac{\lambda}{2n}|x - x_d|^2
\end{cases}.
\end{align}
\noindent The Deep Galerkin Method is sensitive to the scaling of the objective. An objective of low magnitude results in the loss being dominated by the gradient terms which results in a smoothing of the solution and possibly worse convergence. Empirically, it is found that multiplying the objective by $n$ mitigates this effect.
\begin{figure}[!htb]
    \centering
    \includegraphics[width=0.9\linewidth, trim={0 0 0 1.5cm}, clip]{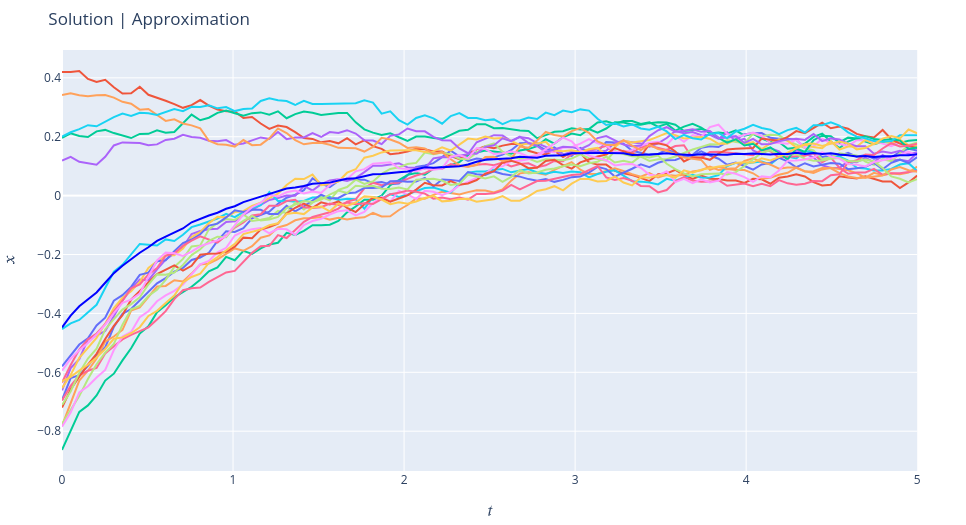}
    \includegraphics[width=0.9\linewidth, trim={0 0 0 1.5cm}, clip]{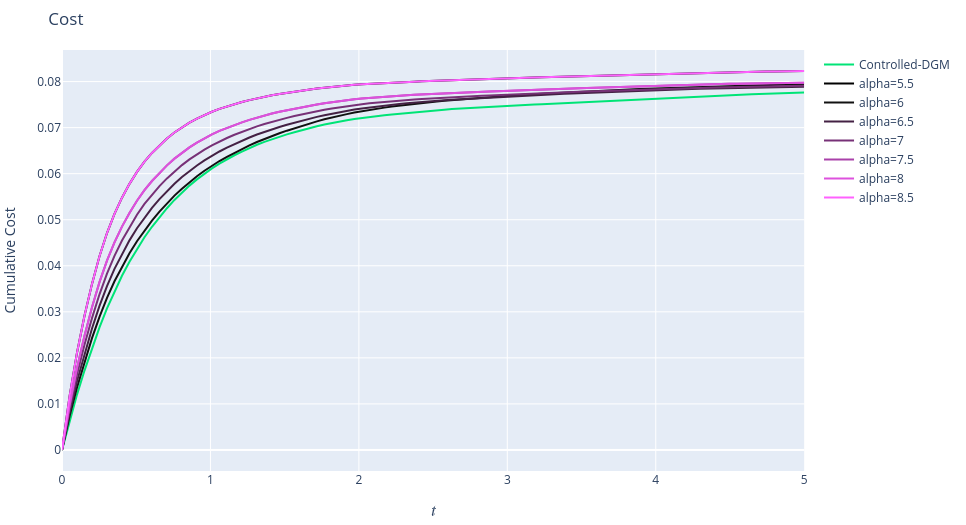}
    \caption{Controlled opinion dynamics in accordance with the Sznajd model and $20$ agents. The interval $[0, 5]$ is discretised with $100$ time points. The problem is specified with $\beta=-3$, $\sigma=0.01$, $\gamma=0.04$, and $\lambda=1$. The lower graph show the associated cumulative cost in comparison to the policy $ \upsilon^{(\alpha)}(x) = \alpha (x_d - x)$. The optimal value lies between $6$ and $8$, however, the DGM-policy performs better in every case. Each agent is associated with one coloured line. The bold blue line in the upper graph represents the empirical average.}
    \label{fig:controlled_sznajd}
\end{figure}
\noindent Fix the time interval at $[0, 5]$ with $\beta=-3$, $\sigma=0.01$, $\gamma=0.04$, and $\lambda=1$. Consider the $20$-agent case. The initial and terminal samplers are implementing Algorithm \ref{alg:clustered_sampling} with a standard deviation of $3.5$ on the truncated Normal. Set $\epsilon$ as a uniform random variable on $(0, 1]$. The terminal sampler includes an explicit reference to the target measure in each batch. The target is set $x_d = 0.2 \cdot \vec{1}$. On the domain, a path sampler as in Algorithm \ref{alg:controlled_path} is used. Interestingly, the policy found by DGM does not contract all sample paths to their exact target but remains slightly below the target. This approach leads to lower cumulative cost than a policy that contracts towards the exact target such as by processes controlled by the policy $\upsilon^{(\alpha)}(x) = \alpha (x_d - x), \alpha > 0$.

\subsection{Hegselmann-Krause Model}
\begin{figure}[!htb]
    \centering
    \includegraphics[width=0.9\linewidth, trim={0 0 0 1.5cm}, clip]{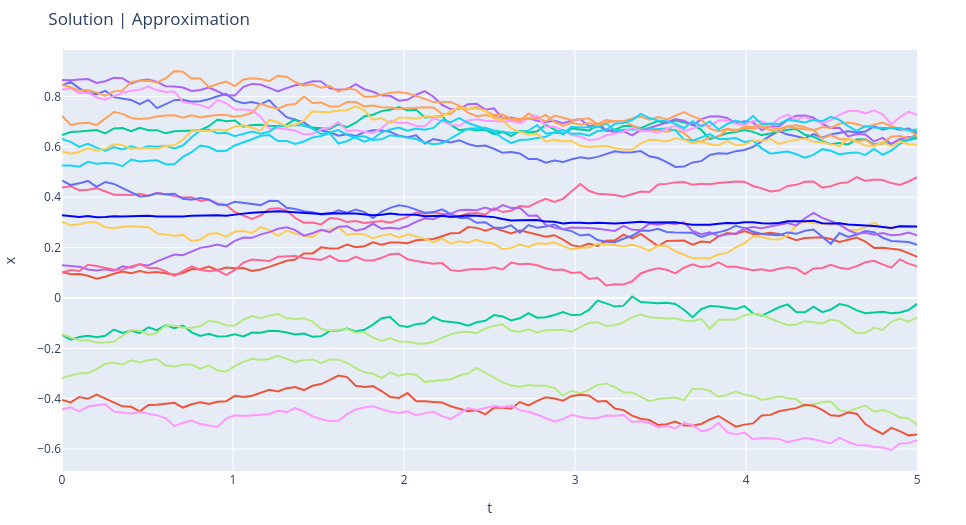}
    \caption{Uncontrolled model in accordance with the Hegselmann-Krause model and $20$ agents. The interval $[0, 5]$ is discretised with $100$ time points. The problem is specified with $\beta=9$, $\kappa=0.2$, $\sigma=0.01$, $\gamma=0.05$, and $\lambda=1$. The measure target is $x_d=0.0$. Each agent is associated with one coloured line. The bold blue line represents the empirical average.}
    \label{fig:uncontrolled_point_hegsel}
\end{figure}
Consider the model by \cite{hegselmann}. The model is similar to that of \cite{sznajd} but differs in the sense that agents exclusively interact with each other in a predefined neighbourhood of radius $\kappa$. It is also distinct in that it considers the flocking of agents, now. To wit, agents clot together as soon as they come close enough. Here, the interaction kernel is scaled up by $\beta>0$. The scalar $\beta$ does not appear in the original model. It is added to accelerate the flocking process such that the effect is observable within a reasonable time frame $[0, T]$. It makes the dynamics more sensitive. This phenomenon can be observed in Figure \ref{fig:uncontrolled_point_hegsel}. Given $\psi(c) = \gamma |c|^2$ and $P(x, y) = \beta \mathbbm{1}_{\{|x-y| \leq \kappa \}}(y)$, the problem is expressed as
\begin{align}
\begin{array}{rll}
     \underset{\{u_i\}_{i \geq 1} \subseteq \mathcal{A}}{\min} & \mathbb{E}^\mathbb{P}\Big[ \int_t^T \frac{1}{2n} \sum_{i=1}^n \lambda |X^{(i)}_s - x_d|^2 \\
     & \hspace{4mm} + \gamma |\upsilon_i(s, X_s)|^2 ds + \frac{\lambda}{2n} ||X_T - x_d||_2^2 \; \Big| \; X_t = x\Big] &\\
     \text{s.t.} & dX^{(i)}_s = \frac{\beta}{n} \sum_{j=1}^n \mathbbm{1}_{\{|X^{(i)}_s-X^{(j)}_s| \leq \kappa \}}(X^{(j)}_s) (X_s^{(j)} \\ & \hspace{12mm} - X_s^{(i)}) ds + \upsilon_i(s, X_s) ds + \sqrt{2 \sigma} dW^{(i)}_s, \\ & \hspace{12mm} \text{ for } 1 \leq i \leq n \\
     & X_0 \sim \nu.
\end{array}
\label{eq:problem_hegsel}
\end{align}

\noindent Fix the time interval at $[0, 5]$ with $\beta=9$, $\sigma=0.01$, $\gamma=0.05$, $\kappa=0.2$, and $\lambda=1$. Consider the $20$-agent case. The initial and terminal samplers are implementing Algorithm \ref{alg:clustered_sampling} with a standard deviation of $3.5$ on the truncated Normal. Set $\epsilon$ as a uniform random variable on $(0, 1]$. The terminal sampler includes an explicit reference of the target measure in each batch. Again, the target is set at $x_d = 0.0 \; \vec{1}$. On the domain, a path sampler as per Algorithm \ref{alg:controlled_path} is used. The setup is very similar to that of the Sznajd model. The internal Euler-Maruyama scheme uses a discretisation of $100$ time points. Its learning rate is $8 \times 10^{-4}$ and the model is trained for $1600$ iterations. The terminal loss is weighted five times higher than the domain loss. The approximation is parameterised by a Residual Neural Network with $370 \; 968$ trainable parameters totalling $1.42$ MB. As for $\alpha$-controlled policies. They seem to perform poorly as soon as a certain number of agents enter each other's interaction radius.
\begin{figure}[!htb]
    \centering
    \includegraphics[width=0.9\linewidth, trim={0 0 0 1.5cm}, clip]{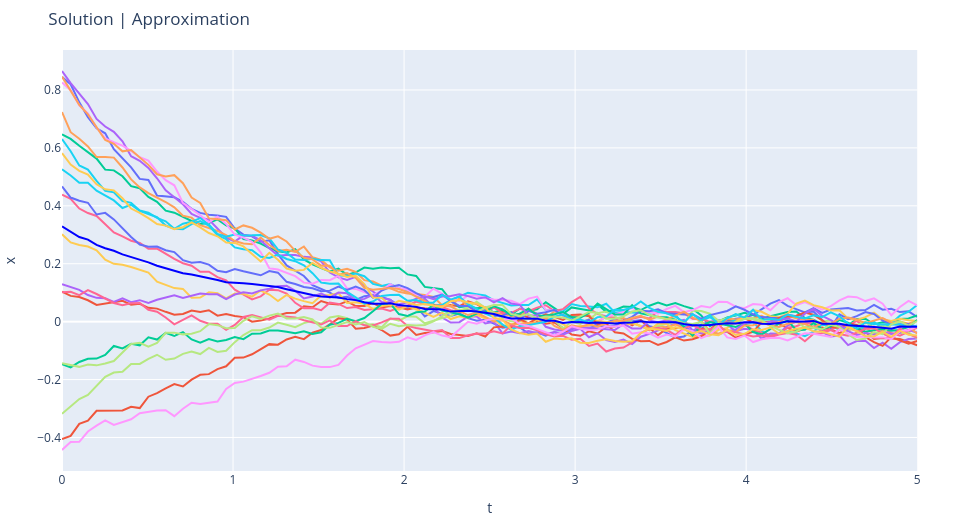}
    \caption{Controlled model in accordance with the Hegselmann-Krause model and $20$ agents. The interval $[0, 5]$ is discretised with $100$ time points. The problem is specified with $\beta=9$, $\kappa=0.2$, $\sigma=0.01$, $\gamma=0.05$, and $\lambda=1$. The measure target is $x_d=0.0$. Each agent is associated with one coloured line. The bold blue line represents the empirical average.}
    \label{fig:point_hegsel}
\end{figure}

\begin{align}
\begin{cases}
    0 = \partial_t J + \frac{ \lambda}{2n}||x-x_d||^2 \\
    \hspace{5mm}+ \frac{\beta}{n} \big(\mathbbm{I}_{\{|x \vec{1}^T - \vec{1} x^T| \preceq K \}}(x) \odot (\vec{1} x^T - x \vec{1}^T) \vec{1}\big)^T \nabla_x J  \\
    \hspace{5mm}+ \sigma \Delta_x J - \frac{n}{2\gamma} ||\nabla_x J||_2^2 \\
    J(T, x) = \frac{\lambda}{2n} ||x - x_d||^2
\end{cases}.
\end{align}

\begin{figure}[!htb]
    \centering
    \includegraphics[width=0.9\linewidth, trim={0 0 0 1.5cm}, clip]{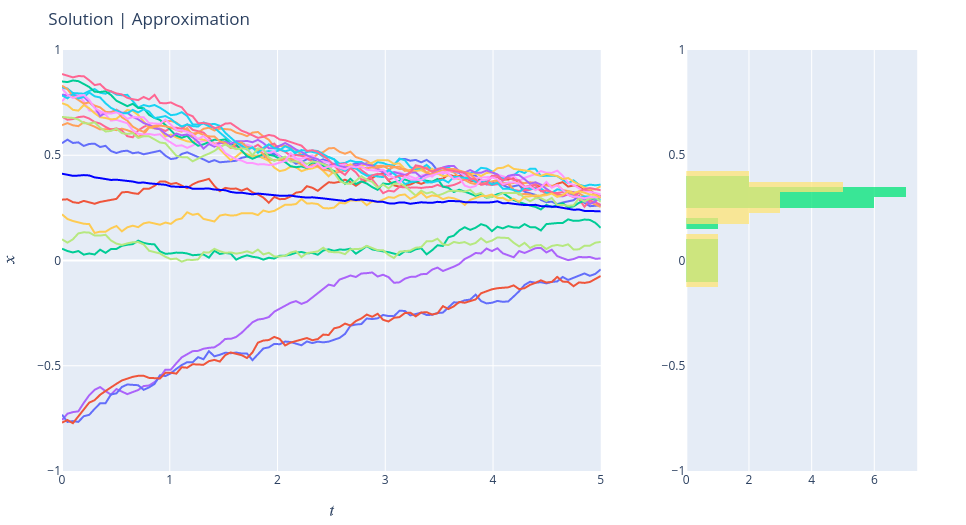}
    \caption{Controlled model in accordance with the Hegselmann-Krause model and $20$ agents. The interval $[0, 5]$ is discretised with $100$ time points. The problem is specified with $\beta=7$, $\kappa=0.2$, $\sigma=0.01$, $\gamma=0.04$, and $\lambda=1$. The target measure included the point masses $0.39, 0.38, 0.37, 0.36, 0.355, 0.34, 0.345, 0.325, 0.32$, $0.29, 0.25, 0.24, 0.23, 0.19, 0.18, 0.1, 0.05, 0.0, -0.05, -0.1$. The horizontal axis of the right figure shows the number of agents. The ground truth is shown in yellow. The approximation is green. Each agent is associated with one coloured line. The bold blue line represents the empirical average.}
    \label{fig:measure_hegsel}
\end{figure}

In the more general sense and for arbitrary target measure $x_d$, consider
\begin{align}
\begin{cases}
    0 = \partial_t J + \frac{ \lambda}{2n}\mathcal{W}^2_{\Omega, 2}(x, x_d) \\
    \hspace{5mm} + \frac{\beta}{n} \big(\mathbbm{I}_{\{|x \vec{1}^T - \vec{1} x^T| \preceq K \}}(x) \odot (\vec{1} x^T - x \vec{1}^T) \vec{1}\big)^T \nabla_x J  \\
    \hspace{5mm}+ \sigma \Delta_x J - \frac{n}{2\gamma} ||\nabla_x J||_2^2 \\
    J(T, x) = \frac{\lambda}{2n} \mathcal{W}^2_{\Omega, 2}(x, x_d)
\end{cases}.
\end{align}
\noindent As an example, $x_d$ is chosen to be an asymmetrical distribution with a long tail. The results are shown in Figure \ref{fig:measure_hegsel}. The model configuration is the same as before, but with a slightly lower $\beta$-value. While the distribution is matched quite well, one can clearly observe that the flocking of the agents produces a steeper distribution. The flocking is strong enough that it appears to be cost-wise more efficient to obtain a slightly altered target distribution. As expected, the ordering from top to bottom is largely maintained for every agent. 
\section{CONCLUSIONS}
 The advantages of a Deep Galerkin approach with Algorithm \ref{alg:controlled_path} can be summarised in three points. Firstly, it is less restrictive in terms of sampling, i.e. it is less dependent on a specific time discretisation. Similarly, one can easily implement any kind of boundary condition on the value function. There are no restrictions on the order of the derivatives as is the case with FBSDE-integration. For high-dimensional HJB PDEs, these methods are far superior to traditional schemes in terms of complexity. Lastly, one can observe a significant improvement on the Deep FBSDE scheme. Hamilton-Jacobi-Bellman PDEs require more carefully chosen sampling. When sampling from the measure of the diffusion process, it is recommended to introduce the control term gradually as its variance is very high in the beginning. Specifically, for interacting diffusion processes, it is not sufficient to retrieve initial samples uniformly. Incorrect sampling has effects on the error convergence and the existence of a solution. The controlled Sznajd and Hegselmann-Krause models can be approached via a Deep Galerkin scheme. 
 The sampling algorithm can certainly be improved upon. However, the challenge persists to keep the respective algorithm at a reasonable complexity. 

\section{Acknowledgements}
This is a preprint version of the paper. It is intended to be uploaded to ResearchGate. The project was supervised by Dr Dante Kalise and Prof Grigorios Pavliotis.
\bibliographystyle{plain}
\bibliography{biblio.bib}

\begin{thebibliography}{10}

\bibitem{al-aradi}
Ali Al-Aradi, Adolfo Correia, Danilo de~Frietas~Naiff, Gabriel Jardim, and Yuri
  Saporito.
\newblock {\em Applications of the Deep Galerkin Method to Solving Partial
  Integro-Differential and Hamilton-Jacobi-Bellman Equations}.
\newblock arXiv preprint arXiv:1912.01455, 2019.

\bibitem{albi}
G.~Albi, L.~Pareschi, and M.~Zanella.
\newblock {\em Boltzmann-Type Control of Opinion Consensus Through Leaders}.
\newblock Phil. Trans. R. Soc. A 372: 20140138, 2014.

\bibitem{kalise}
Giacomo Albi, Young-Pil Choi, Massimo Fornasier, and Dante Kalise.
\newblock {\em Mean Field Control Hierarchy}.
\newblock Springer Science+Business Media, 2017.

\bibitem{albrekht}
E.~Al'Brekht.
\newblock {\em On the Optimal Stabilization of Nonlinear Systems}.
\newblock J. Appl. Math. Mech, 1961.

\bibitem{alla}
A.~Alla, M.~Falcone, and S.~Volkswein.
\newblock {\em Error Analysis for POD Approximations of Infinite Horizon
  Problems via the Dynamic Programming Approach}.
\newblock SIAM J. Control Optim., 2017.

\bibitem{bokanowski}
O.~Bokanowski, J.~Garckea, M.~Griebel, and I.~Klompmaker.
\newblock {\em An Adaptive Sparse Grid Semi-Lagrangian Scheme for First Order
  Hamilton-Jacobi-Bellman Equations}.
\newblock J. Sci. Comput., 2013.

\bibitem{degond}
Pierre Degond, Michael Herty, and Jian-Guo Liu.
\newblock {\em Meanfield Games and Model Predictive Control}.
\newblock arXiv preprint arXiv:1412.7517v1, 2014.

\bibitem{dolgov}
Sergey Dolgov, Dante Kalise, and Karl~K. Kunisch.
\newblock {\em Tensor Decomposition Methods for High-Dimensional
  Hamilton-Jacobi-Bellman Equations}.
\newblock Society for Industrial and Applied Mathematics, 2019.

\bibitem{weinan-e2}
Weinan E, Jiequn Han, and Arnulf Jentzen.
\newblock {\em Deep Learning-Based Numerical Methods for High-Dimensional
  Parabolic Partial Differential Equations and Backward Stochastic Differential
  Equations}.
\newblock arXiv preprint arXiv:1706.04702v1, 2017.

\bibitem{weinan-e}
Weinan E and Bing Yu.
\newblock {\em The Deep Ritz Method: A Deep Learning-Based Numerical Algorithm
  for Solving Variational Problems}.
\newblock Springer-Verlag GmbH Germany, 2018.

\bibitem{exarchos}
Ioannis Exarchos and Evangelos~A. Theodorou.
\newblock {\em Stochastic optimal control via forward and backward stochastic
  differential equations and importance sampling}.
\newblock ScienceDirect, 2017.

\bibitem{falcone}
M.~Falcone and R.~Ferretti.
\newblock {\em Semi-Lagrangian Approximation Schemes for Linear and
  Hamilton-Jacobi Equations}.
\newblock Society for Industrial and Applied Mathematics, 2013.

\bibitem{hegselmann}
Rainer Hegselmann and Ulrich Krause.
\newblock {\em Opinion Dynamics and Bounded Confidence Models, Analysis, and
  Simulation}.
\newblock Journal of Artifical Societies and Social Simulation (JASSS) vol.5,
  no. 3, 2002.

\bibitem{kalise2}
Dante Kalise and Axel Kröner.
\newblock {\em Reduced-Order Minimum Time Control of
  Advection-Reaction-Diffusion Systems via Dynamic Programming}.
\newblock HAL Id: hal-01089887, 2014.

\bibitem{kalman}
R.~E. Kalman.
\newblock {\em Contributions to the Theory of Optimal Control}.
\newblock Bol. Soc. Mat. Mexicana, 1960.

\bibitem{kang}
W.~Kang and L.~C. Wilcox.
\newblock {\em Mitigating the Curse of Dimensionality: Sparse Grid
  Characteristics Method for Optimal Feedback Control and HJB Equations}.
\newblock Comput. Optim. Appl., 2017.

\bibitem{munos}
R.~Munos, L.C. Baird, and A.W. Moore.
\newblock Gradient descent approaches to neural-net-based solutions of the
  hamilton-jacobi-bellman equation.
\newblock In {\em IEEE/IJCNN'99. International Joint Conference on Neural
  Networks. Proceedings (Cat. No.99CH36339)}, volume~3, pages 2152--2157 vol.3,
  1999.

\bibitem{nguyen}
Thanh Nguyen, Binh Pham, Trung~T. Nguyen, and Binh~T. Nguyen.
\newblock A deep learning approach for solving poisson’s equations.
\newblock In {\em IEEE/12th International Conference on Knowledge and Systems
  Engineering (KSE)}, pages 213--218, 2020.

\bibitem{raissi}
Maziar Raissi.
\newblock {\em Forward-Backward Stochastic Neural Networks: Deep Learning of
  High-dimensional Partial Differential Equations}.
\newblock arXiv preprint arXiv:1804.07010, 2018.

\bibitem{sirignano}
Justin Sirignano and Konstantinos Spiliopoulos.
\newblock {\em DGM: A deep learning algorithm for solving partial differential
  equations}.
\newblock ScienceDirect, 2018.

\bibitem{stefansson}
Elis Stefansson and Yoke~Peng Leong.
\newblock Sequential alternating least squares for solving high dimensional
  linear hamilton-jacobi-bellman equation.
\newblock In {\em IEEE/RSJ International Conference on Intelligent Robots and
  Systems (IROS)}, pages 3757--3764, 2016.

\bibitem{sznajd}
Katarzyna Sznajd-Weron and Józef Sznajd.
\newblock {\em Opinion Evolution in Closed Community}.
\newblock International Journal of Modern Physics C VOL. 11, NO. 06, 2000.

\end{thebibliography}

\pagebreak

\onecolumn
\appendix

\subsection{Proof of Theorem \ref{th:bound}}
Let $\delta_\theta(t, x): \Omega \times [0, \infty) \mapsto \mathbb{R}$ denote the difference of the parameterised to the true value function. We note that for a process as in Equation \ref{eq:agent_dynamics}, Itô's Chain Rule gives

\begin{align*}
    & \int_t^T (\partial_s + \mathcal{L}^{u})\delta_\theta ds = \delta_\theta(T, X_T) -  \delta_\theta(t, X_t) - \sigma \int_t^T \nabla \cdot \delta_\theta dW_s \\
    \Leftrightarrow & \mathbb{E}^\mathbb{P}\Big[\int_t^T (\partial_s + \mathcal{L}^{u})\delta_\theta ds \Big]^2 = \mathbb{E}^\mathbb{P}[\delta_\theta(T, X_T) -  \delta_\theta(t, X_t)]^2 + \sigma^2 \mathbb{E}^\mathbb{P}\Big[\int_t^T \nabla \cdot \delta_\theta dW_s \Big]^2 \\
    \Leftrightarrow & \big|\big|\int_t^T (\partial_s + \mathcal{L}^{u})\delta_\theta ds \big|\big|^2_{L^2(\Omega; \mathbb{P})} = ||\delta_\theta(T, X_T) -  \delta_\theta(t, \cdot)||^2_{L^2(\Omega; \mathbb{P})} + \sigma^2 \int_t^T \big|\big| \nabla \delta_\theta(t, \cdot) \big|\big|^2_{L^2(\Omega; \mathbb{P})} ds \\
    & \hspace{40.6mm} \geq ||\delta_\theta(T, X_T) -  \delta_\theta(t, \cdot)||^2_{L^2(\Omega; \mathbb{P})} \\
    \Leftrightarrow & \big|\big|\int_t^T (\partial_s + \mathcal{L}^{u})\delta_\theta ds \big|\big|_{L^2(\Omega; \mathbb{P})} \geq ||\delta_\theta(t, \cdot)||_{L^2(\Omega; \mathbb{P})} - ||\delta_\theta(T, X_T)||_{L^2(\Omega; \mathbb{P})}
\end{align*}
Now, applying the Cauchy-Schwartz Inequality on the left-hand side yields:
\begin{align*}
     & (T-t)^{\frac{1}{2}} \Big( \int_\Omega \int_t^T \big| (\partial_s + \mathcal{L}^{u})\delta_\theta \big|^2 ds \; d \mathbb{P}\Big)^{\frac{1}{2}} \geq ||\delta_\theta(t, \cdot)||_{L^2(\Omega; \mathbb{P})} - ||\delta_\theta(T, X_T)||_{L^2(\Omega; \mathbb{P})} \\
    \Leftrightarrow & \sqrt{T-t} \big|\big|(\partial_s + \mathcal{L}^{u})\delta_\theta \big|\big|_{L^2([t, T] \times \Omega; \mathbb{P})} + ||\delta_\theta(T, X_T)||_{L^2(\Omega; \mathbb{P})} \geq ||\delta_\theta(t, \cdot)||_{L^2(\Omega; \mathbb{P})} \\
    \Leftrightarrow & \sqrt{T-t} \big|\big|(\partial_t + \mathcal{L}^u) J_\theta + F(\cdot, \cdot, u) \big|\big|_{L^2([t, T] \times \Omega; \mathbb{P})} + ||J_\theta(T, X_T) - G(X_T)||_{L^2(\Omega; \mathbb{P})} \geq ||J_\theta(t, \cdot) - J(t, \cdot)||_{L^2(\Omega; \mathbb{P})} 
\end{align*}

\subsection{Proof of Theorem \ref{th:bound_2}}
Let $(\Omega, \mathcal{F}, \mathbb{P})$ be a probability space and $\sigma_{\nu_1}(Z_t)$ be the sigma algebra generated by the $\nu_1$-random variable $Z_t$. Let $Y_t$ be well-defined on $\sigma_{\nu_1}(Z_t)$. Further, let the running cost and terminal cost be bounded by below with $F\geq B_F \in \mathbb{R}^+$ and $G\geq B_G \in \mathbb{R}^+$. Let $M$ denote the set of $L^2(\Omega; \mathbb{P})$ random variables measurable with respect to $\sigma_{\nu_1}(Z_t)$. $M$ is a closed linear subspace of $L^2(\Omega; \mathbb{P})$. The minimisation is, therefore, given by the projection of $Y_t$ onto $M$. By the projection theorem, we have
\begin{align*}
    Y_t = \mathbb{E}^\mathbb{P}[Y_t | X_t \in \sigma_{\nu_1}(Z_t)] + (Y_t - \mathbb{E}^\mathbb{P}[Y_t | X_t \in \sigma_{\nu_1}(Z_t)])
\end{align*}
and the minimal norm is achieved with 
\begin{align*}
    J_\theta(X_t) = \mathbb{E}^\mathbb{P}[Y_t | X_t \in \sigma_{\nu_1}(Z_t)].
\end{align*}
We assume that we can achieve an ideal approximation. The resulting norm is
\begin{align}
     \nonumber \min_\theta||Y_t-J_\theta(Z_t)||^2_{L^2(\Omega; \mathbb{P})} &= ||Y_t-\mathbb{E}^\mathbb{P}[Y_t | X_t \in \sigma_{\nu_1}(Z_t)]||^2_{L^2(\Omega; \mathbb{P})} \\
    \nonumber &= \Big|\Big|\mathbb{E}^{\mathbb{P}}\Big[ \int_t^T F(s, X_s, u_s) ds + G(X_T) \;\Big|\; X_t \in \mathcal{F}_t \Big] \\
    & \hspace{6mm} -\mathbb{E}^\mathbb{P}\Big[\mathbb{E}^{\mathbb{P}}\Big[ \int_t^T F(s, Z_s, u_s) ds + G(Z_T) \;\Big|\; Z_t = X_t \Big] \; \Big| \; X_t \in \sigma_{\nu_1}(Z_t)\Big]\Big|\Big|^2_{L^2(\Omega; \mathbb{P})} \\ \nonumber
    &= \Big|\Big|\mathbb{E}^{\mathbb{P}}\Big[ \int_t^T F(s, X_s, u_s) ds + G(X_T) \;\Big|\; X_t \in \mathcal{F}_t \setminus \sigma_{\nu_1}(Z_t) \Big]\Big|\Big|^2_{L^2(\Omega; \mathbb{P})} \\ \nonumber
    &= \int_{\Omega} \frac{1}{(\mathbb{P}( X_t \in \mathcal{F}_t \setminus \sigma_{\nu_1}(Z_t)))^2} \Big|\int_{\mathcal{F}_t \setminus \sigma_{\nu_1}(Z_t)} \int_t^T F(s, X_s, u_s) ds + G(X_T) \Big|^2 d\mathbb{P} \\ \nonumber
    &\geq \int_{\Omega} \Big|\int_{X_t \in  \mathcal{F}_t \setminus \sigma_{\nu_1}(Z_t)} \int_t^T F(s, X_s, u_s) ds + G(X_T) d\mathbb{P} \Big|^2 d\mathbb{P} \\ \nonumber
    &\geq \int_{\Omega} \Big|\int_{\mathcal{F}_t \setminus \sigma_{\nu_1}(Z_t)} (T-t) B_F + B_G d\mathbb{P} \Big|^2 d\mathbb{P} \\
    &= ((T-t) B_F + B_G)^2 ||\mathbb{P}(X_t \in  \mathcal{F}_t \setminus \sigma_{\nu_1}(Z_t))||^2_{L^2(\Omega; \mathbb{P})}.
\end{align}

\end{document}